\documentclass[10pt,twocolumn,letterpaper]{article}

\usepackage{iccv}
\usepackage{times}
\usepackage{epsfig}
\usepackage{graphicx}
\usepackage{amsmath}
\usepackage{amssymb}
\usepackage{mathtools}

\usepackage{multirow}
\usepackage{bm}

\newcommand{\squishlist}{
	\begin{list}{$\bullet$}
		{ \setlength{\itemsep}{0pt}
			\setlength{\parsep}{1pt}
			\setlength{\topsep}{1pt}
			\setlength{\partopsep}{0pt}
			\setlength{\leftmargin}{1.5em}
			\setlength{\labelwidth}{1em}
			\setlength{\labelsep}{0.5em} } }
\newcommand{\squishend}{\end{list} 
}

\usepackage[toc,page]{appendix}

\usepackage[pagebackref=true,breaklinks=true,letterpaper=true,colorlinks,bookmarks=false]{hyperref}

\iccvfinalcopy % *** Uncomment this line for the final submission

\def\iccvPaperID{2888} % *** Enter the ICCV Paper ID here

\ificcvfinal\fi

\begin{document}

%%%%%%%%% TITLE
\title{Learning to Generate Scene Graph from Natural Language Supervision}

\author{
Yiwu Zhong\textsuperscript{1}, Jing Shi\textsuperscript{2}, Jianwei Yang\textsuperscript{3}, Chenliang Xu\textsuperscript{2}, Yin Li\textsuperscript{1}\\
\textsuperscript{1}University of Wisconsin-Madison \quad 
\textsuperscript{2}University of Rochester \quad
\textsuperscript{3}Microsoft Research\\
{\tt\small \{yzhong52,yin.li\}@wisc.edu} \quad 
{\tt\small \{j.shi,chenliang.xu\}@rochester.edu} \quad
{\tt\small jianwei.yang@microsoft.com}
}
% {\tt\small yzhong52@wisc.edu} \quad 
% {\tt\small j.shi@rochester.edu} \quad
% {\tt\small jianwei.yang@microsoft.com} \\
% {\tt\small chenliang.xu@rochester.edu} \quad
% {\tt\small yin.li@wisc.edu} \quad

\maketitle
\ificcvfinal\thispagestyle{empty}\fi

%%%%%%%%% ABSTRACT
\begin{abstract}
Learning from image-text data has demonstrated recent success for many recognition tasks, yet is currently limited to visual features or individual visual concepts such as objects. In this paper, we propose one of the first methods that learn from image-sentence pairs to extract a graphical representation of localized objects and their relationships within an image, known as scene graph. To bridge the gap between images and texts, we leverage an off-the-shelf object detector to identify and localize object instances, match labels of detected regions to concepts parsed from captions, and thus create ``pseudo'' labels for learning scene graph. Further, we design a Transformer-based model to predict these ``pseudo'' labels via a masked token prediction task. Learning from only image-sentence pairs, our model achieves 30\% relative gain over a latest method trained with human-annotated unlocalized scene graphs. Our model also shows strong results for weakly and fully supervised scene graph generation. In addition, we explore an open-vocabulary setting for detecting scene graphs, and present the first result for open-set scene graph generation. Our code is available at \url{https://github.com/YiwuZhong/SGG_from_NLS}.
\end{abstract}

%%%%%%%%% BODY TEXT
%-------------------------------------------------------------------------
% introduction
\section{Introduction}

\begin{figure}[t]
	\centering
	\includegraphics[width=0.99\linewidth]{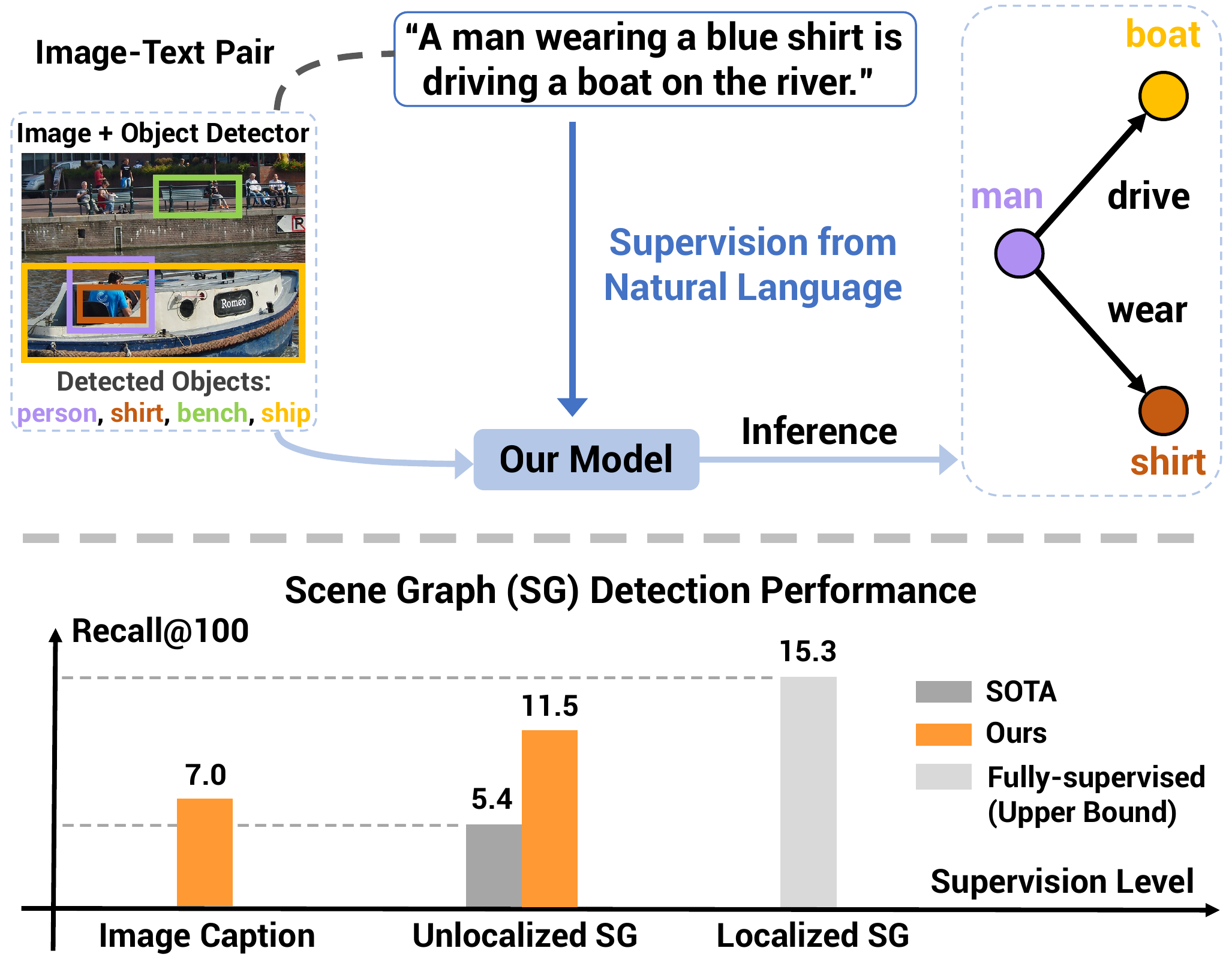}
    \caption{{\bf Top (our setting)}: Our goal is learning to generate localized scene graphs from image-text pairs. Once trained, our model takes an image and its detected objects as inputs and outputs the image scene graph. {\bf Bottom (our results)}: A comparison of results from our method and~\cite{zareian2020weakly} with varying levels of supervision.}
    \label{fig:teaser}%\vspace{-1.0em}
\end{figure}

% background: learning from natural language supervision
An image might have millions of pixels, yet its visual content can be often summarized using dozens of words. Images and their text descriptions (\ie captions) are available in great abundance from the Internet~\cite{sharma2018conceptual}, and offer a unique opportunity of image understanding aided by natural language. Learning visual knowledge from image-text pairs has been a long-standing problem~\cite{wang2009learning,chen2013neil,divvala2014learning,chen2015webly,joulin2016learning,yang2018visual,ye2019cap2det,gupta2020contrastive,zareian2020open,sariyildiz2020learning,desai2020virtex,radford2021learning}, with recent success on learning deep models for visual representation~\cite{chen2015webly,joulin2016learning,sariyildiz2020learning,desai2020virtex,radford2021learning}, and for recognizing and detecting individual visual concepts (\eg objects)~\cite{yang2018visual,ye2019cap2det,zareian2020open,radford2021learning,gupta2020contrastive}. In this paper, we ask the question: {\it can we learn to detect visual relationships beyond individual concepts from image-text pairs?} Fig.\ \ref{fig:teaser} (a) illustrates an example of such relationships (``man {\it drive} boat'').

% scene graph and stoa
As a first step, we focus on learning scene graph generation (SGG) from image-sentence pairs. A scene graph is a symbolic and graphical representation of an image, with each graph node as a localized object and each edge as a relationship (\eg a predicate) between a pair of objects. Scene graph has emerged as a structured representation for many vision tasks, including action recognition~\cite{ji2020action}, 3D scene understanding~\cite{armeni20193d,wald2020learning}, image generation and editing~\cite{johnson2018image,dhamo2020semantic}, and vision-language tasks (\eg image captioning~\cite{yang2019auto,yao2018exploring,zhong2020comprehensive} and visual question answering~\cite{shi2019explainable,teney2017graph,hudson2019learning}). Most previous scene graph methods~\cite{xu2017scene,li2017scene,zellers2018neural,li2018factorizable,yang2018graph,chen2019counterfactual, tang2019learning,tang2020unbiased,zhang2019graphical} follow a fully supervised approach, relying on human annotations of object bounding boxes, object categories and their relationships. These annotations are very costly and difficult to scale. Recently, Zareian \etal \cite{zareian2020weakly} considered weakly supervised learning of scene graphs from image-level labels of unlocalized scene graphs. Nonetheless, learning scene graphs from images and their text descriptions remains unexplored.

% key challenge and why using object detector
A major challenge of learning scene graphs from image-sentence pairs is the missing link between many candidate image regions and a few concepts (\eg nouns and predicates) parsed from an image caption. To this end, we propose to leverage off-the-shelf object detectors, capable of identifying and localizing object instances from hundreds of common categories. Our key idea is that object labels of detected image regions can be further matched to sentence concepts, and thus provide ``pseudo'' labels for learning scene graphs, thereby bridging the gap between region-concept pairs. Our hypothesis is that these ``pseudo'' labels, coupled with a large-scale dataset, can be used for training a deep model to detect scene graph of an input image. Our language supervised setting is shown in Fig.\ \ref{fig:teaser} (a).

% our method
Inspired by the recent success of vision-language pretraining~\cite{chen2020uniter,li2020oscar,zhou2020unified,Lu_2020_CVPR,Su2020VL-BERT,tan2019lxmert,lu2019vilbert}, we develop a Transformer-based model for learning to generate scene graphs supervised by image-sentence pairs. Specifically, our model takes inputs of visual features from a pair of detected object regions, text embeddings of their predicted categorical labels, and contextual features from other object regions, all provided by an off-the-shelf detector~\cite{faster-rcnn}. Our model then learns to recognize the visual relationship between the input object pair, represented as a localized subject-predicate-object (SPO) triplet. A scene graph can thus be generated by enumerating all pairs from a small set of detected objects. During training, our model learns from only image-sentence pairs using ``pseudo'' labels produced by matching the detected object labels to the parsed sentence concepts. During inference, our model generates a scene graph given an input image with its detection results. 

% our results
Our model is trained on captioning datasets including COCO Caption~\cite{chen2015microsoft} and Conceptual Caption~\cite{sharma2018conceptual}, and evaluated on Visual Genome~\cite{krishna2017visual} --- a widely used scene graph benchmark. Our results, summarized in Fig.\ \ref{fig:teaser} (b), significantly outperform the state of the art~\cite{zareian2020weakly} on weakly supervised SGG by a relative margin of $30\%$, despite that our model only requires image-sentence pairs for training while \cite{zareian2020weakly} is trained using human-annotated unlocalized scene graphs. With the same supervision as~\cite{zareian2020weakly}, our model achieves a relative gain of $112\%$ in recall. Further, our model also demonstrates strong results on fully supervised SGG. While these results are reported on closed-set setting with known target concepts during training, we also present promising results on open-set SGG where the concept vocabulary is crafted from image captions. Our work is among the first methods for learning to detect scene graphs from only image-sentence pairs, and presents the first result for open-set SGG. We believe our work provides a solid step towards structured image understanding. 

This paper was accepted to ICCV 2021. In this arXiv version, we add additional comparison to a concurrent work from~\cite{ye2021linguistic}, provide more details during the discussion of our results, and include additional results in the appendix.  

%-------------------------------------------------------------------------
% related work
\section{Related Work}
We briefly review recent works on learning visual knowledge from natural language and scene graph generation, with a focus on the development of deep models. 

\noindent \textbf{Learning Visual Knowledge from Language}. The availability of images and their text descriptions on the Internet has spurred a surge of interest in learning from image-text pairs. Early works focused on learning from image-hashtag pairs for visual representation learning~\cite{chen2015webly,joulin2016learning} and for recognizing objects, scenes, and actions~\cite{chen2013neil,divvala2014learning,farhadi2009describing, lampert2009learning}. More recent works have shifted attention to learning from images and their sentence descriptions. For example, image-sentence pairs were used for visual representation learning via image captioning~\cite{desai2020virtex}, image-text matching~\cite{radford2021learning}, or image-conditioned language modeling~\cite{sariyildiz2020learning}, and for visual and textual representation learning using context prediction tasks~\cite{chen2020uniter,li2020oscar,zhou2020unified,Lu_2020_CVPR,Su2020VL-BERT,tan2019lxmert,lu2019vilbert}. Image captions were exploited for object recognition~\cite{wang2009learning} and object detection~\cite{ye2019cap2det,jerbi2020learning, yang2018visual,zareian2020open}. Unlike these previous works, our work learns to detect localized scene graphs that encode objects and their relationships in an input image. Inspired by previous works on visual and textual representation learning, we propose a Transformer-based model for scene graph generation and formulate the problem as predicting masked tokens of subject, predicate and object.

\noindent \textbf{Fully Supervised Scene Graph Generation}. An image scene graph represents localized object instances as nodes and their relationships as edges on the graph. Scene graph generation (SGG) aims to extract this graphical representation from an input image. A related problem is visual relationship detection (VRD)~\cite{yu2017visual, lu2016visual,zhang2017visual,dai2017detecting} that also localizes objects and recognizes their relationships yet without the notation of a graph. Thanks to the development of large-scale densely annotated image scene graph datasets, such as Visual Genome (VG) dataset~\cite{krishna2017visual}, a large array of methods have been proposed for scene graph generation. Several different models have been explored, including iterative message passing~\cite{xu2017scene, li2017scene}, recurrent network~\cite{zellers2018neural}, tree structure encoding~\cite{tang2019learning,wang2020sketching}, graph convolution and pruning~\cite{li2018factorizable,yang2018graph}, casual reasoning~\cite{chen2019counterfactual,tang2020unbiased} and contrastive learning~\cite{zhang2019graphical}. A major drawback of these approaches is the requirement of human-annotated, localized scene graphs with categorical labels and locations of all nodes and edges. Our work seeks to address this drawback by learning to detect scene graphs from only image-sentence pairs.

\noindent \textbf{Weakly-supervised Scene Graph Generation}. Several recent works have explored weakly supervised settings for VRD~\cite{peyre2017weakly,baldassarre2020explanation,zhang2017ppr} and SGG~\cite{zhang2017ppr,zareian2020weakly,Shi_ICCV_2021}. Most of them addressed the task of VRD and seeks to learn from unlocalized SPO triplets. For example, Peyre \etal \cite{peyre2017weakly} proposed to assign image-level labels to pairs of detected objects via discriminative clustering. Baldassarre \etal \cite{baldassarre2020explanation} first predicted visual predicates given the detected objects, and then retrieved the subjects and objects using backward explanation techniques. Zhang \etal \cite{zhang2017ppr} designed a fully convolutional network to jointly learn object detection and predicate prediction from image-level labels, using object proposals as model inputs. They reported results on both VRD and SGG. The most relevant work is from Zareian \etal \cite{zareian2020weakly}. They proposed learning from unlocalized scene graphs for SGG, and developed a message passing mechanism to update features of detected objects and to gradually refine labels of objects and predicates. Our recent work \cite{Shi_ICCV_2021} presented a simple baseline for weakly supervised SGG using first-order graph matching. Similar to these approaches, our method explores learning using less labels for SGG. Unlike previous approaches, our method leverages image captions --- a different type of labels that are easier to obtain than unlocalized SPO triplets or scene graphs. A concurrent work from Ye \etal \cite{ye2021linguistic} also explored learning scene graph from image-sentence pairs. They proposed to use visual grounding to iteratively match the detected image regions and the text entities parsed from captions. Unlike their method, we leverage an object detector to create the pseudo labels for SPO triplets, leading to significantly better empirical results. Our work is thus among the first methods to detect scene graphs by learning from only image-sentence pairs.

%-------------------------------------------------------------------------
% method
\section{Scene Graph from Language Supervision}
\label{method}
\begin{figure*}
	\centering
	\includegraphics[width=0.9\linewidth]{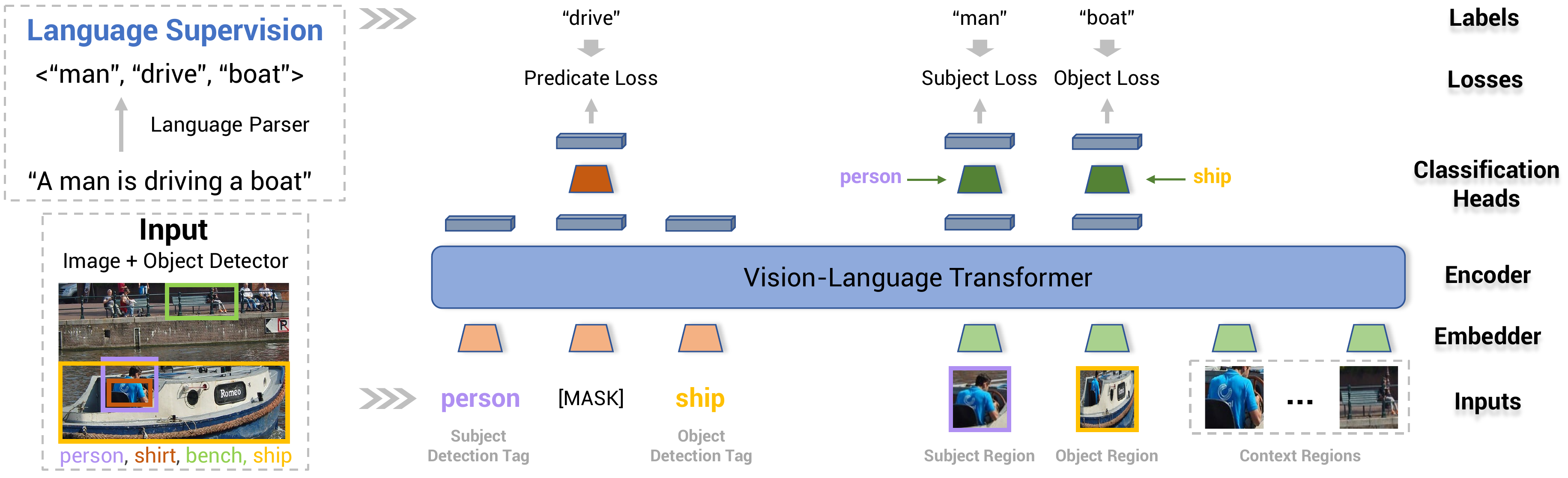}
    \caption{Overview of our proposed model for language supervised scene graph generation. Given an image, an object detector is first applied with the detected objects as the inputs to our model. Our model further embeds the detected region features and textual object categories (e.g., the tags of a pair of subject-object, the MASK representing the predicate) into token embeddings, followed by a multi-layer Transformer encoder. Finally, our model predicts the labels of the subject region, the object region and the predicate.}
    \label{fig:model_overview}
\end{figure*}

% problem formulation: learning from language supervision
With a large collection of paired images $\{I\}$ and captions $\{S\}$, our goal is learning to detect an image scene graph $\mathcal{G}=(\mathcal{V},\mathcal{E})$ from an input image $I$. $\mathcal{G}$ is a directed graph with nodes $\mathcal{V}$ and edges $\mathcal{E}$. Each node $v_i \in \mathcal{V}$ denotes a localized object in $I$, represented by its bounding box $b_i$ and object label $o_i$ within a vocabulary $\mathcal{C}^g_o$. Each edge $e_{ij} \in \mathcal{E}$ denotes a predicate (\eg ``drive'') from a vocabulary $\mathcal{C}^g_p$ pointing from node $v_i$ to node $v_j$, where $T_{ij} = (v_i, e_{ij}, v_j)$ defines a triplet of subject-predict-object (SPO). Scene graph generation is thus a challenging structured output prediction problem.

% object detector -> scene graph generation
Similar to previous SGG methods~\cite{xu2017scene,zhang2017visual,tang2019learning,zellers2018neural,zareian2020weakly}, we assume a set of object regions $R = \{r_n\}$ provided by an detector. Each region $r_n= (\bar{b}_n, \bar{o}_n)$ consists of a bounding box $\bar{b}_n$ and a predicted object category $\bar{o}_n$ from a vocabulary $\mathcal{C}^d_o$ given by the detector. $r_n$ thus defines a candidate node of the target scene graph $\mathcal{G}$. It is worth noting that the vocabulary of the detector $\mathcal{C}^d_o$ is different from the vocabulary of the scene graph $\mathcal{C}^g_o$ (\ie $\mathcal{C}^d_o \neq \mathcal{C}^g_o$). With object regions $R = \{r_n\}$, SGG is reduced to classify $r_n$ into object categories ($\mathcal{C}^g_o \cup \{\text{background}\}$), and infer the predicate label ($\mathcal{C}^g_p \cup \{\text{background}\}$) between each subject-object region pair $(r_k, r_l)$. A main innovation of our model is to learn from only image-sentence pairs for SGG, without the need of ground-truth object labels nor their relationships. 

% overview of our method: matching; labeling of triplet; 
\noindent \textbf{Learning from Language Supervision}. Our key idea is to extract SPO triplets from an image caption, and match these triplets to object categories of image regions given by the detector, thereby creating ``pseudo'' labels for these regions and their relationships. Specifically, we adopt a language parser to extract a set of triplets $\{T'\}$ from the caption $S$. We further link object region pairs $\{r_k, r_l\}$ in the image $I$ provided by the detector to the parsed sentence triplets $T'$. This is done by matching detected object categories $\bar{o}_k$ and $\bar{o}_l$ from every region pair to the subject and object in each $T'$, respectively. If matched, the sentence triplet $T'$ will define a ``pseudo'' label for the region pair $(r_k, r_l)$ (subject, object) and their relationship $e_{kl}$ (predicate). These ``pseudo'' labels can then be used to train our model.

\begin{table}[]
\centering
\resizebox{0.45\textwidth}{!}{%
\begin{tabular}{c|c|c|c}
\hline
\multirow{3}{*}{\begin{tabular}[c]{@{}c@{}}Scene Graph\\ Generation\\ Settings\end{tabular}} & \multicolumn{3}{c}{Required Annotation during Training} \\ \cline{2-4} 
 & \multirow{2}{*}{\begin{tabular}[c]{@{}c@{}}Image \\ Description\end{tabular}} & \multirow{2}{*}{\begin{tabular}[c]{@{}c@{}}Object\&Predicate \\ Category Labels\end{tabular}} & \multirow{2}{*}{\begin{tabular}[c]{@{}c@{}}Object \\ Boxes\end{tabular}} \\
 &  &  &  \\ \hline
Fully Supervised~\cite{xu2017scene} &  & \checkmark & \checkmark \\
Weakly Supervised~\cite{zareian2020weakly} &  & \checkmark &  \\
Language Supervised (ours) & \checkmark &  &  \\ \hline
\end{tabular}}\vspace{0.3em}
\caption{Our language supervised setting vs.\ fully supervised and weakly supervised settings. Our method learns from only image-sentence pairs to generate localized image scene graphs.}
\label{tab:setting} 
\end{table}

\noindent \textbf{Comparison to Fully and Weakly Supervised Settings}. Our setting of learning to generate scene graphs from image-sentence pairs is different from previous fully and weakly supervised settings, as shown in Table~\ref{tab:setting}. Our setting provides a new opportunity of learning structured visual knowledge from natural language supervision.

\noindent \textbf{Overview of Our Model}. Our model, inspired by recent works in vision-language pretraining~\cite{chen2020uniter,li2020oscar,zhou2020unified,Lu_2020_CVPR,Su2020VL-BERT,tan2019lxmert,lu2019vilbert}, seeks to label the SPO triplet given a pair of regions. Specifically, we design a Transformer-based model with its inputs as a region pair $(r_k, r_l)$ and the contextual features from other regions $\{{r_n}\}-\{{r}_k, {r}_l\}$. Our model then predicts the category labels $(o_k, e_{kl}, o_l)$ of a SPO triplet $T_{kl}$ for the input region pair $(r_k, r_l)$. During training, our model is supervised by the ``pseudo'' labels $T'$ parsed from caption $S$. During inference, our model takes inputs of the image $I$ and its detection results $R=\{r_n\}$, labels every region pair $(r_k, r_l)$, and aggregates the SPO triplets into a full scene graph. Fig.\ \ref{fig:model_overview} illustrates our model.

\subsection{Triplet Transformer}\label{section:transformer_model}
Our proposed Triplet Transformer is a triplet labeling model based on an input region pair and its contextual features. Specifically, for each region $r_n = (\bar{b}_n, \bar{o}_n)$, we denote its visual, positional, and textual features as $\mathbf{x}_n^r$, $\mathbf{x}_n^p$, and $\mathbf{x}_n^o$, respectively. $\mathbf{x}_n^r$ is the visual feature (ROI) pooled from the region $\bar{b}_n$. $\mathbf{x}_n^p$ is a feature encoding the position of the bounding box, \ie, a 7-D vector with the normalized top/left/bottom/right coordinates, width, height and area for the region box $\bar{b}_n$. $\mathbf{x}_n^o$ is the word embedding of the region object label $\bar{o}_n$. Given an input region pair $(r_k, r_l)$ and all other detected regions, our model builds a composition function $f=g\circ h$ to predict the labels $(o_k, e_{kl}, o_l)$ of a SPO triplet, given by
\begin{equation*}
\resizebox{.48\textwidth}{!}{
$o_k, e_{kl}, o_l = g\circ h\left(\underbrace{\mathbf{x}_k^o, \mathbf{x}_l^o;}_\text{Textual Embedder} 
\underbrace{\mathbf{x}_k^r, \mathbf{x}_l^r, \mathbf{x}_k^p, \mathbf{x}_l^p; \quad
\overbrace{\{\mathbf{x}_u^r, \mathbf{x}_u^p\}_{u\neq k, l}}^\text{ Contextual Features} }_\text{Visual Embedder}\right)$
}
\end{equation*}
where $u$ indexes all regions except $r_k$ and $r_l$.

Our model thus consists of: (1) a visual embedder that encodes visual and positional region features; (2) a textual embedder that embeds textual region features (from object labels); (3) a multi-layer Transformer $h$ that conducts message passing among the input visual and textual embeddings; and (4) classification heads $g$ that predict the labels of a triplet. We now present details for each component.

\noindent \textbf{Visual Embedder}. Our visual embedder transforms visual and positional features ($\mathbf{x}_n^r$ and $\mathbf{x}_n^p$) of region $r_n$ into an embedding $\mathbf{v}_n$, where $n$ indexes all region features including $k$ (subject), $l$ (object) and $u$ (context). This is given by 
\begin{equation}
\small
\mathbf{v}_n = {\rm LN}({\rm LN}(\mathbf{W}_r \mathbf{x}_n^r) + {\rm LN}(\mathbf{W}_p \mathbf{x}_n^p) + \mathbf{e}_n^t),
\end{equation}
where $\mathbf{W}_r$ and $\mathbf{W}_p$ are trainable weights that project the features into the same dimension $d$. $\mathbf{e}_n^t \in \mathbb{R}^d$ is the type embedding of a region (subject vs.\ object vs.\ context). $\rm LN$ denotes Layer Normalization~\cite{ba2016layer}.

\noindent \textbf{Textual Embedder}. Our textual embedder accepts two inputs: (1) the word embeddings $\mathbf{x}^o_k$ and $\mathbf{x}^o_l$ of region labels  for subject and object region, respectively; and (2) the word embedding of a special word ``MASK'', denoted as $\mathbf{x}^o_p$, representing the missing predicate. The embedder encodes the input word embedding and the positional embedding into a textual embedding $\mathbf{t}_m$, given by
\begin{equation}
\small
\mathbf{t}_m = {\rm LN}(\mathbf{W}_e \mathbf{x}_m^o + \mathbf{e}_m^p),
\end{equation}
where $m$ indexes $k$ (subject), $p$ (predicate) or $l$ (object). $\mathbf{e}_m^p \in \mathbb{R}^d$ is the positional embedding~\cite{devlin-etal-2019-bert} of the current token. $\mathbf W_e$ represents the trainable weights projecting the word embedding into the dimension of $d$.

\noindent \textbf{Transformer Encoder}. The visual and textual embeddings ($\mathbf{v}_n$ and $\mathbf{t}_m$) are further fed into a multi-layer Transformer encoder~\cite{transformer}. This encoder uses multi-head self-attention, coupled with multilayer perceptron (MLP) and layer normalization, to output a contextualized embedding ($\hat{\mathbf{v}}_n \in \mathbb{R}^d$ or $\hat{\mathbf{t}}_m \in \mathbb{R}^d$) for each input $\mathbf{v}_n$ or $\mathbf{t}_m$. This Transformer encoder can be considered as conducting message passing across all input tokens. Among all the outputs, the embeddings corresponding to the subject, predicate, object tokens will be further used for triplet label prediction, as shown in Fig.\ \ref{fig:model_overview}. For a region pair $(r_k, r_l)$, the embeddings  $\hat{\mathbf{v}}_k$ / $\hat{\mathbf{t}}_k$ correspond to the visual / textual feature of the subject region (\ie the first input region), the predicate embedding $\hat{\mathbf{t}}_p$ is from the special word ``MASK'', and the embeddings $\hat{\mathbf{v}}_l$ / $\hat{\mathbf{t}}_l$ represent the visual / textual feature of the object region (\ie the second input region). 

\noindent \textbf{Classification Heads}. Our model further fuses the encoder outputs, and predicts labels of a SPO triplet (subject-predicate-object) for the input region pair $(r_k, r_l)$. The feature fusion is given by
\begin{equation}\small
\begin{split}
& \mathbf s = \hat{\mathbf{v}}_k + \mathbf W_v \mathbf{x}_k^o,\ \ \ \ \mathbf o = \hat{\mathbf{v}}_l + \mathbf W_v \mathbf{x}_l^o,\\
& \mathbf p = \hat{\mathbf{t}}_p + \mathbf W_{ts} \hat{\mathbf{t}}_k + \mathbf W_{to} \hat{\mathbf{t}}_l + \mathbf W_{vs} \hat{\mathbf{v}}_k + \mathbf W_{vo} \hat{\mathbf{v}}_l,
\end{split}
\end{equation} 
where $\mathbf W_v$, $\mathbf W_{ts}$, $\mathbf W_{to}$, $\mathbf W_{vs}$, $\mathbf W_{vo}$ are learnable weights. The outputs $\mathbf{s} \in \mathbb{R}^d$, $\mathbf{o} \in \mathbb{R}^d$, $\mathbf{p} \in \mathbb{R}^d$ are further used to classify subject, predicate, and object labels, respectively. This is done using a two-layer MLP followed by softmax.

\subsection{Learning from Language Supervision}
\label{section:training}
Our key innovation is the use of image captions as the only supervisory signal for training our model. This is done by constructing ``pseudo'' labels of triplets from image captions. Concretely, we first parse a caption into a set of SPO triplets. Each triplet is further matched to every pair of regions, by comparing subject and object tokens in the sentence triplet to the predicted categories of a region pair. Our model is then trained on the matched pairs of regions to predict their corresponding sentence triplets. We point out that our approach of learning from image-sentence pairs can be easily adapted by different SGG models. 

\noindent \textbf{Closed-Set vs.\ Open-Set}. In this paper, we primarily consider a closed-set setting --- the vocabulary of the subject, predicate, and object during evaluation is known in prior. In this setting, our learning is focused on the concepts of interest and our model only considers sentence triplets within the vocabulary. Nonetheless, our method does support the open-set setting, where there are no limits on the vocabulary. In this case, our model learns from all frequently appearing subject, predicate, and object tokens in the captions. Additional matching step is needed at inference time to identify concepts in the target vocabulary. We will explore this setting in our experiment.  

\noindent \textbf{Triplet Parsing and Filtering}. We use an off-the-shelf rule-based language parser~\cite{sgparser} based on Schuster \etal \cite{schuster-etal-2015-generating} to parse the triplets in the image captions. After parsing, the triplets with the lemmatized words for subject, predicate and object are obtained. We further perform an optional filtering step on the initial collection of triplets. For the closed-set setting, we only keep concepts that can be matched to the categories in the target vocabulary. Two concepts are matched if (1) there is overlapping between their synsets, lemmas or hypernyms in WordNet~\cite{miller1995wordnet} (\eg ``tortoise'' $\rightarrow$ ``animal''), or (2) if their root forms can be matched (\eg ``baseball player'' $\rightarrow$ ``player'').

\noindent \textbf{Pseudo Label Assignment}. 
With the filtered triplets, our next step is to match sentence triplets to pairs of regions provided by the object detector. This is done by a greedy matching between every triplet from the caption and each region pair from the image. Specifically, we match the corresponding subject and object tokens between a triplet and a region pair, again using a token's synsets, the synsets' lemmas and hypernyms in WordNet~\cite{miller1995wordnet} and its root form. If multiple triplets are matched to the same region pair, we randomly select one of them. We also filter out region pairs that does not overlap and far away from each other, following~\cite{zellers2018neural}, as these pairs are less likely to contain a relationship. Once matched, the triplet is considered as the pseudo label of the region pair for training our model.

\noindent \textbf{Model Training}. Our model is trained by predicting the pseudo labels of the region pairs. We apply a multi-class cross-entropy loss for the subject, predicate, and object, respectively. Our final loss function is given by
\begin{equation}
    \small
    \mathcal{L} =  \lambda_s \mathcal{L}_s +  \lambda_p \mathcal{L}_p + \lambda_o  \mathcal{L}_o
\end{equation}
where $\mathcal{L}_s$, $\mathcal{L}_p$, and $\mathcal{L}_o$ is the loss for subject, predicate, and object respectively. And $\lambda_s$, $\lambda_p$, and $\lambda_o$ are their corresponding loss weights. We set $\lambda_s=\lambda_o=0.5$ and $\lambda_p=1$ following previous work~\cite{zellers2018neural,zareian2020weakly}.

\noindent \textbf{Weighted Loss}. One challenge for learning is the domain gap between (a) image-sentence pairs used for training and (b) images and their target scene graphs during inference. For example, the distributions of concepts might be quite different in image-sentence pairs vs.\ image scene graphs. In the closed-set setting, we might have an estimated frequency of the concepts on scene graphs. In this case, we apply a weighted loss during training, where the weight for each category is set to the ratio between the frequency of the token in image-sentence pairs and the estimated frequency of the matched tokens in scene graphs. If a category is not matched to any target category, no loss weight will be applied. This weighted loss function only requires an estimated frequency of concepts on the target dataset, and can be considered as a simple approach for domain adaption.

\noindent \textbf{Model Inference}.
Once trained, our model takes a region pair and its contextual features, and predicts a SPO triplet. To obtain a scene graph, we enumerate all possible region pairs and feed them into our model. The predicted probabilities are further averaged for each region and thus each region is predicted to single category. In the open-set setting, an additional matching step is needed to infer the probability of target categories based on the predicted categories from image-sentence pairs. In this case, we apply the same matching step in our label assignment step.

\noindent \textbf{Extension to Weakly and Fully Supervised Settings}. 
Our model can be easily extended to weakly and fully supervised settings. In weakly supervised setting, we replace triplets parsed from captions with those from unlocalized scene graphs~\cite{zareian2020weakly}, and follow the same label assignment of our setting. For fully supervised setting, we simply replace our pseudo labels with ground-truth scene graph labels.

%-------------------------------------------------------------------------
% experiments
\section{Experiments and Results}
\label{section:experiments}

\begin{table*}[t]
\centering
\resizebox{0.75\textwidth}{!}{%
\begin{tabular}{c|c|c|c|c|c|cc}
\hline
\multirow{2}{*}{Method} & \multicolumn{5}{c|}{Training Setting} & \multicolumn{2}{c}{SGDet} \\ 
 & Supervision & Level & Source & \#Triplets & \#Images & R@50 & R@100 \\ \hline
Ours+Full & Localized Scene Graph & Full & Visual Genome & 406K & 58K & \textbf{13.8} & \textbf{15.3} \\ \hline \hline
VSPNet~\cite{zareian2020weakly} & \multirow{3}{*}{Unlocalized Scene Graph} & \multirow{3}{*}{Weak} & \multirow{3}{*}{Visual Genome} & \multirow{3}{*}{406K} & \multirow{3}{*}{58K} & 4.7 & 5.4 \\
VSPNet$\dagger$ &  &  &  &  &  & 6.7 & 7.4 \\ 
Ours+Weak &  &  &  &  &  & \textbf{10.0} & \textbf{11.5} \\ \hline 
Ours+MotifNet & \multirow{2}{*}{Image Description}  & \multirow{2}{*}{Weaker} & \multirow{2}{*}{CC + COCO} & \multirow{2}{*}{313K} & \multirow{2}{*}{210K} & 5.6 & 6.7 \\
Ours &  &  &  &  &  & \textbf{5.9} & \textbf{7.0} \\ \hline 
\end{tabular}}\vspace{0.3em}
\caption{Results of language supervised SGG. Different from all previous approaches, our model can learn from image-sentence pairs for SGG. With only image-sentence pairs as the supervisory signal, our model outperforms VSPNet --- a latest method of weakly supervised SGG trained using human-annotated and unlocalized scene graphs.}
\label{tab:weakly-SGG}
\end{table*}

\begin{table}[t]
\centering
\resizebox{0.48\textwidth}{!}{%
\begin{tabular}{c|c|c|cc}
\hline
\multirow{2}{*}{Method} & \multicolumn{2}{c|}{Training Setting} & \multicolumn{2}{c}{SGDet} \\ 
 & Supervision & Source & R@50 & R@100 \\ \hline
LSWS\cite{ye2021linguistic} & \multirow{2}{*}{Unlocalized Scene Graph} & \multirow{2}{*}{Visual Genome} & 7.3 & 8.7 \\
Ours &  &  & \textbf{10.0} & \textbf{11.5} \\ \hline 
LSWS\cite{ye2021linguistic} & \multirow{4}{*}{Image Description} & Visual Genome & 3.9 & 4.0 \\
Ours &  & Visual Genome  & \textbf{9.2} & \textbf{10.3} \\
LSWS\cite{ye2021linguistic} &  & COCO  & 3.3 & 3.7   \\
Ours &  & COCO  & \textbf{5.8} & \textbf{6.7} \\ \hline 
%Ours &  & CC + COCO  & \textbf{5.9} & \textbf{7.0} \\ \hline 
\end{tabular}}\vspace{0.3em}
\caption{Comparison to the concurrent work of LSWS~\cite{ye2021linguistic}.}
\label{tab:concurrent}
\end{table}

We now present our experiments and results. We start with our main results on learning SGG from image-sentence pairs, followed by our ablation studies. Further, we present results on fully supervised SGG, and explore open-set SGG.

\noindent \textbf{Datasets}. To evaluate our model, we used the standard split~\cite{xu2017scene} of Visual Genome (VG)~\cite{krishna2017visual} (150 objects, 50 predicates, 75K/32K images for train/test). VG comes with human-annotated image captions and localized scene graphs, and is a widely used benchmark for SGG. We also considered image captions on VG for our ablation study, and localized scene graphs on VG for fully supervised SGG. 

For training, we considered image captions from VG, COCO Caption (COCO)~\cite{chen2015microsoft}, and Conceptual Caption (CC)~\cite{sharma2018conceptual}. COCO contains 123K images with each labeled by 5 human-annotated captions. We selected 106k images in COCO for training by filtering out images that exist in the test set of VG. CC contains 3.3M image-caption pairs automatically collected from alt-text enabled images on the web. For the closed-set setting where the target categories are known, we matched the parsed tokens from each dataset to target categories, and kept 148-52, 143-56, 148-64 object-predicate categories for VG, COCO and CC, respectively, leading to 673K/75K (triplets/images) on VG, 154K/64K on COCO, and 159K/145K on CC.

\noindent \textbf{Evaluation Protocol and Metrics}. For the majority of our experiments, we evaluate Scene Graph Detection (SGDet) following the protocol from~\cite{xu2017scene}. SGDet captures both the localization and classification performance using metrics of Recall@K (R@K) \cite{lu2016visual,xu2017scene} and mean Recall@K (mR@K) \cite{chen2019knowledge,tang2019learning}. R@K computes the recall between the top K predicted triplets and ground-truth ones. A predicted triplet is considered as correct only when all requirements are met: (1) the predicted triplet labels match one of the ground-truth triplet, (2) the detected subject-object regions match the ground-truth subject-object regions with an IoU $\ge$ 0.5, respectively. mR@K averages R@K across all predicate categories. We also included Scene Graph Classification (SGCls) and Predicate Classification (PredCls) in our experiment on fully SGG. Importantly, all experiments were conducted with graph constraint that limits each subject-object pair to have only one predicate prediction.

\noindent \textbf{Implementation Details}. We used a Faster R-CNN~\cite{faster-rcnn} detector pre-trained on OpenImages~\cite{kuznetsova2020open}, capable of detecting 601 object categories. We kept the top 36 objects per image and extracted the 1536-D region features from the detector. The object tags were represented by the 300-D GloVe embeddings~\cite{pennington2014glove}. We adopted the Transformer implementation from UNITER~\cite{chen2020uniter} with 2 self-attention layers, 12 attention heads in each layer and hidden size $d=768$. SGD optimizer was used in training with the image batch of 32, 16 sampled triplets per image, and the initial learning rate of 0.0032. We used the benchmark implementation provided by Tang \etal \cite{tang2020unbiased} for evaluation.

\subsection{Language Supervised Scene Graph Generation}
\label{SGG}
We now present our main results on learning to generate scene graphs from image-sentence pairs.

\noindent \textbf{Setup and Baselines}. We consider several baselines and variants of our model. A key feature of our model is the ability to learn from only image-sentence pairs. 
\squishlist
\item \textbf{VSPNet}~\cite{zareian2020weakly} is designed for weakly supervised SGG and learns from unlocalized scene graph. As our close competitor, VSPNet takes the inputs of object proposals from the same OpenImage detector used by our model.
\item \textbf{VSPNet$\dagger$} further augments VSPNet with object box predictions from the detector. VSPNet$\dagger$ thus has the same input image regions as our model.
\item \textbf{Ours+Weak} is our model trained using unlocalized scene graphs, same as the setting of VSPNet. 
\item \textbf{Ours+MotifNet} combines our pseudo label assignment with a supervised SGG model (MotifNet~\cite{zellers2018neural}). This model is thus trained using only image-sentence pairs.
\item \textbf{Ours+Full} is our model trained with full supervision and using ground-truth scene graph labels. This should be considered as an upper bound of our model.
\squishend

\noindent \textbf{Results}. Table~\ref{tab:weakly-SGG} presents our main results. With image description (CC + COCO) as only supervision, our models (Ours/Ours+MotifNet) significantly outperform VSPNet trained using unlocalized scene graphs (7.0/6.7 vs.\ 5.4 R@100), despite that image-sentence pairs are much weaker supervisory signals. Our Transformer-based model also beats Ours+MotifNet, and performs on par with the improved version of VSPNet (VSPNet$\dagger$) (7.0 vs.\ 7.4 R@100). When trained using unlocalized scene graphs, our model (Ours+Weak) again outperforms VSPNet variants by a large margin (11.5 vs.\ 5.4/7.4 R@100). These results provide convincing evidence that our model can learn from only image-sentence pairs to detect scene graph in an image with high quality. Finally, there is a noticeable gap between Ours and Ours+Weak (7.0 vs.\ 11.5 R@100), and between Ours+Weak and Ours+Full (11.5 vs.\ 15.3 R@100), suggesting ample room for future work.

\begin{figure*}
	\centering
	\includegraphics[width=0.95\linewidth]{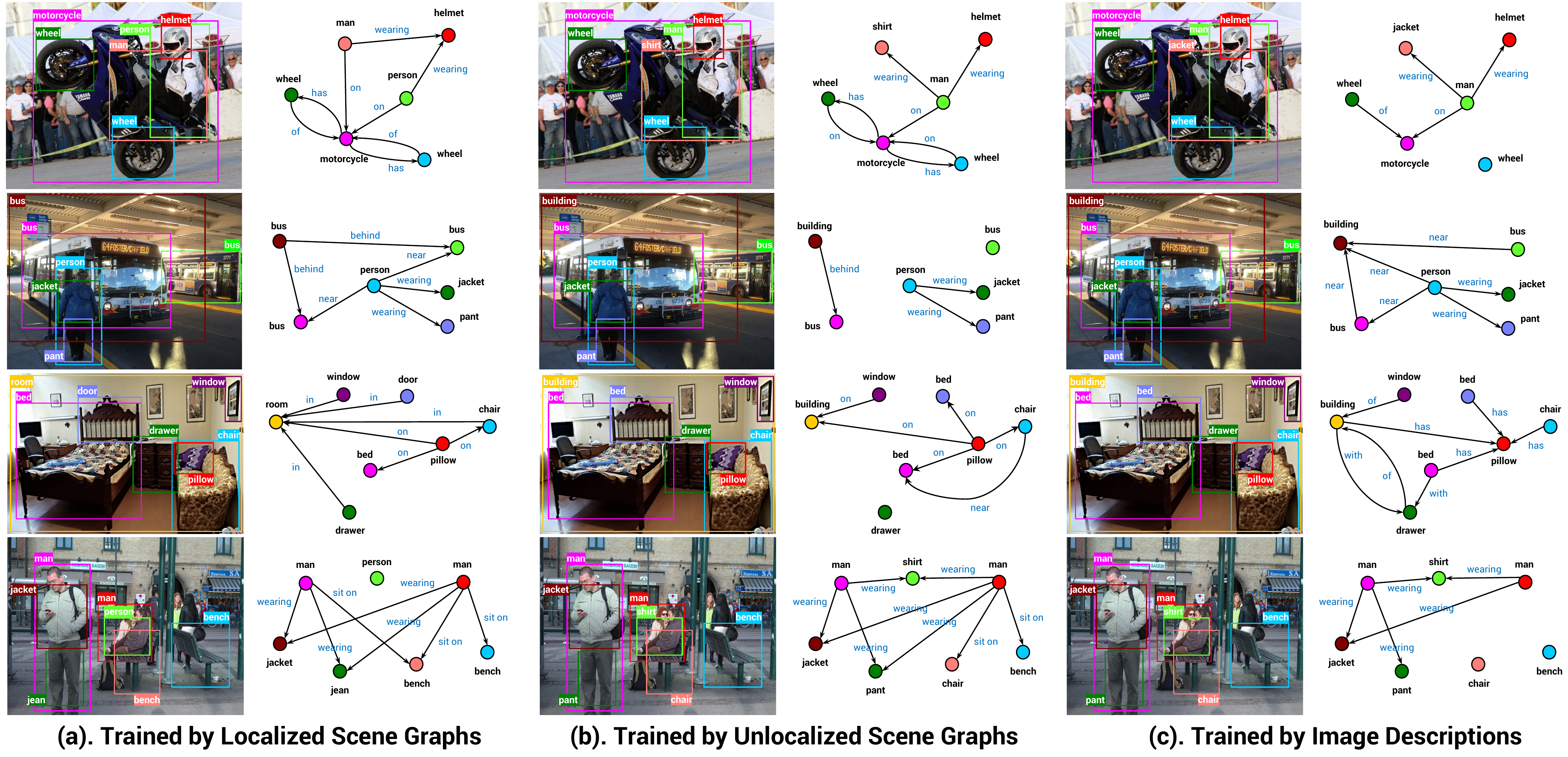}
    \caption{Qualitative results of our models on VG test set for SGG. All models take the same detected regions and predict the scene graph labels. In each row, we show 3 identical images and the corresponding scene graphs generated from the models trained by different levels of supervision. The visualized relationships are picked from the top 30 predicted triplets.}
    \label{fig:sample_vis}
\end{figure*}

Fig.\ \ref{fig:sample_vis} further visualizes the output scene graphs from our models, including Ours+Full (left), Ours+Weak (middle) and Ours (right) in Table~\ref{tab:weakly-SGG}. Our model trained by image-sentence pairs produces scene graphs with a comparable quality as those trained using strong supervision (\eg ``man-on-motorcycle'' and ``man-wearing-helmet'' in the 1st row). Further, our models trained using scene graphs tend to predict a different set of predicate when compared to our model trained using image-sentence pairs. This is best illustrated in the 3rd row of Fig.\ \ref{fig:sample_vis} (``on'' vs.\ ``has''). We conjecture that this is caused by different distributions of predicates in scene graph and in image captions. Finally, it is worth noting that similar to many previous approaches, our models fall short when common sense reasoning is needed. This is shown in the 4th row of Fig.\ \ref{fig:sample_vis}, where our models predict two man wearing the same jacket or shirt.

\noindent \textbf{Comparison to LSWS~\cite{ye2021linguistic}}. In addition, we compare our results to a concurrent work of LSWS~\cite{ye2021linguistic}. LSWS also learns to generate scene graph from image-sentence pairs using iterative visual grounding. Table~\ref{tab:concurrent} summarizes the comparison. When trained with the same level of supervision and the same dataset, our models constantly outperform LSWS by a large margin. For example, when trained using image-sentence pairs on COCO, our method achieves 5.8 R@50 and 6.7 R@100 vs.\ 3.3 R@50, and 3.7 R@100 from LSWS --- a relative gain of at least 75\%. When trained with unlocalized scene graph as the setting in VSPNet~\cite{zareian2020weakly}, our model also outperforms LSWS by a noticeable margin (+2.7 R@50 and +2.8 R@100).

\subsection{Ablation Studies} 
\label{section:ablation}
We now present ablation studies of our method.

% table: ablation study on the sources of image descriptions
\begin{table}[t]
\centering
\resizebox{0.47\textwidth}{!}{%
\begin{tabular}{cccc|c|c|cc}
\hline
\multicolumn{3}{c}{Source of Image Description} & \multirow{2}{*}{\begin{tabular}[c]{@{}c@{}}Weighted\\ Loss\end{tabular}} & \multirow{2}{*}{\#Triplets} & \multirow{2}{*}{\#Images} & \multicolumn{2}{c}{SGDet} \\ \cline{1-3}
CC & COCO & VG &  &  &  & R@50 & R@100 \\ \hline
\checkmark &  &  &  & 159K & 145K & 3.4 & 4.1 \\
 & \checkmark &  &  & 154K & 64K & 3.8 & 4.5 \\ \hline 
\checkmark &  &  & \checkmark & 159K & 145K & 5.3 & 6.4 \\
 & \checkmark &  & \checkmark & 154K & 64K & 5.8 & 6.7 \\ 
\checkmark & \checkmark &  & \checkmark & 313K & 210K & 5.9 & 7.0 \\ \hline \hline
 &  & \checkmark & - & 673K & 75K & 9.2 & 10.3 \\ \hline
\end{tabular}}\vspace{0.3em}
\caption{Ablation study on different sources of image descriptions and weight loss for training our model.}
\label{tab:description-SGG}
\end{table}

\noindent \textbf{Sources of Image Descriptions.} Table~\ref{tab:description-SGG} presents results of our model trained using different sources of captions. Not surprisingly, the model trained on VG performs better (10.3 R@100) than the model trained on CC (4.1 R@100) or COCO (4.5 R@100), since the evaluation dataset is also VG. Interestingly, the model trained on CC performs on par with the model trained on COCO with similar number of triplets, despite that captions on COCO are manually annotated and has higher quality than those harvested from the Internet on CC. We thus conjecture that the performance of our model is minorly influenced by the caption quality.

\noindent \textbf{Effects of Weighted Loss}. Table~\ref{tab:description-SGG} also compares the use of different loss functions. Adding weighted loss improves the recall from 4.1 R@100 to 6.4 R@100 when using CC as the training data. This result indicates that using weighted loss can effectively close the domain gap between datasets. For example, the predicate ``wearing'' appears frequently in VG yet occurs rarely in CC and COCO. With weighted loss, the recall of ``wearing'' is improved by 22.1 R@100.

% table: ablation study on detectors and label assignment
\begin{table}[t]
\centering
\resizebox{0.46\textwidth}{!}{%
\begin{tabular}{c|c|c|cc}
\hline
\multirow{2}{*}{Model} & \multirow{2}{*}{Object Detector} & \multirow{2}{*}{Lable  Assignment} & \multicolumn{2}{c}{SGDet} \\
 &  &  & R@50 & R@100 \\ \hline
VSPNet~\cite{zareian2020weakly} & OpenImages & Iterative Alignment & 4.7 & 5.4 \\
VSPNet$\dagger$ & OpenImages & Iterative Alignment & 6.7 & 7.4 \\
MotifNet & OpenImages & Detection Tags (Ours) & 9.3 & 10.7 \\
Ours & OpenImages & Detection Tags (Ours) & 10.0 & 11.5 \\
Ours & Objects365 & Detection Tags (Ours) & 6.1 & 6.4 \\ \hline
\end{tabular}}\vspace{0.3em}
\caption{Ablation study on object detectors and label assignment schemes. Results are reported using unlocalized scene graphs as supervision. Our proposed label assignment scheme provides consistent performance boost across methods, while the choice of object detectors has a major impact of the performance.}
\label{tab:matching-SGG}
\end{table}

\noindent \textbf{Effects of Label Assignment}. We evaluate our label assignment scheme in Table~\ref{tab:matching-SGG}. Specifically, we consider the weakly supervised setting of learning from unlocalized scene graphs, apply our method to MotifNet~\cite{zellers2018neural}, and present the results in 3rd row. With our scheme, MotifNet beats the latest VSPNet (10.7 vs.\ 7.4 R@100), suggesting the effectiveness of our label assignment scheme.

\noindent \textbf{Effects of Object Detector}. We additionally consider another object detector trained on Object365 dataset~\cite{Shao_2019_ICCV}. In Table~\ref{tab:matching-SGG}, our model with the Objects365 detector has lower recall (6.4 R@100) than our model with the OpenImages detector (11.5 R@100). Upon a close inspection, we conclude that the recall drop is mainly caused by the mismatch between the object categories of the detector and those in VG. Particularly, we find only 94 (out of 150) VG objects can be matched to Objects365 categories while 123 VG objects can be matched to OpenImages categories. For example, ``shirt'' and ``building'' are most frequent objects in VG. Objects365 detector cannot detect them while OpenImages detector can. As a result, the triplets involving these objects will not be used to train the model, and the trained model fails to detect these concepts.

% table: ablation study on text input & visual input branch
\begin{table}[t]
\centering
\resizebox{0.41\textwidth}{!}{%
\begin{tabular}{cc|c|cc}
\hline
\multicolumn{2}{c|}{Model Inputs} & \multirow{2}{*}{\begin{tabular}[c]{@{}c@{}}Object Detection\\ mAP\end{tabular}} & \multicolumn{2}{c}{SGDet} \\ \cline{1-2}
Text Input & Visual Input &  & R@50 & R@100 \\ \hline
\checkmark & \checkmark & 10.7 & 10.0 & 11.5 \\ 
           & \checkmark & 10.6 & 3.9 & 4.7 \\
\checkmark &            & 6.9 & 6.2 & 7.7 \\ \hline
\end{tabular}}\vspace{0.3em}
\caption{Ablation study on different model inputs. Results are reported using unlocalized scene graphs as supervision. Visual and textual features complement to each other for SGG.}
\label{tab:model-SGG}
\end{table}

% table: fully-supervised setting
\begin{table*}[t]
\centering
\resizebox{0.96\textwidth}{!}{%
\begin{tabular}{c|ccc|ccc|ccc|ccc|ccc|ccc}
\hline
\multirow{3}{*}{Model} & \multicolumn{9}{c|}{Recall} & \multicolumn{9}{c}{Mean Recall} \\ \cline{2-19} 
 & \multicolumn{3}{c|}{SGDet} & \multicolumn{3}{c|}{SGCls} & \multicolumn{3}{c|}{PredCls} & \multicolumn{3}{c|}{SGDet} & \multicolumn{3}{c|}{SGCls} & \multicolumn{3}{c}{PredCls} \\
 & @20 & @50 & @100 & @20 & @50 & @100 & @20 & @50 & @100 & @20 & @50 & @100 & @20 & @50 & @100 & @20 & @50 & @100 \\ \hline
IMP~\cite{xu2017scene} & 18.1 & 25.9 & 31.2 & 34.0 & 37.5 & 38.5 & 54.3 & 61.1 & 63.1 & 2.8 & 4.2 & 5.3 & 5.2 & 6.2 & 6.5 & 8.9 & 11.0 & 11.8 \\
VTransE~\cite{zhang2017visual} & 23.0 & 29.7 & 34.3 & 35.4 & 38.6 & 39.4 & 59.0 & 65.7 & 67.6 & 3.7 & 5.0 & 6.0 & 6.7 & 8.2 & 8.7 & 11.6 & 14.7 & 15.8 \\
VCTree~\cite{tang2019learning} & 24.7 & 31.5 & 36.2 & \textbf{37.0} & \textbf{40.5} & \textbf{41.4} & \textbf{59.8} & \textbf{66.2} & \textbf{68.1} & 4.2 & 5.7 & 6.9 & 6.2 & 7.5 & 7.9 & 11.7 & 14.9 & 16.1 \\
MotifNet~\cite{zellers2018neural} & \textbf{25.1} & \textbf{32.1} & \textbf{36.9} & 35.8 & 39.1 & 39.9 & 59.5 & 66.0 & 67.9 & 4.1 & 5.5 & 6.8 & 6.5 & 8.0 & 8.5 & 11.5 & 14.6 & 15.8 \\
Ours & 24.6 & 31.8 & 36.3 & 36.5 & 40.0 & 40.8 & 58.7 & 65.6 & 67.4 & \textbf{5.3} & \textbf{7.3} & \textbf{8.7} & \textbf{8.3} & \textbf{10.4} & \textbf{11.1} & \textbf{13.3} & \textbf{17.7} & \textbf{19.5} \\ \hline
\end{tabular}}\vspace{0.3em}
\caption{Results of fully supervised SGG. All models use the same object detector pre-trained on the VG dataset, and the same codebase provided by Tang et al.\ \cite{tang2020unbiased} for evaluation. Results of previous models come from Tang et al.\ \cite{tang2020unbiased}.}
\label{tab:supervised-SGG}
\end{table*}

\noindent \textbf{Textual vs.\ Visual Inputs}. Finally, We study the contribution of model inputs. This is done by probing our trained model during inference and masking out one input at a time. For textual input, we substitute subject and object embeddings for a random vector. For visual input, we replace the original region features with the averaged region features in the current image. The results are presented in Table~\ref{tab:model-SGG}. Using only visual input leads to a minor drop in detection mAP (10.6 vs.\ 10.7) and a large drop in scene graph recall (4.7 vs.\ 11.5 R@100), indicating a major performance drop in predicate prediction. In contrast, using only text input has a large drop mAP (6.9 vs.\ 10.7) and a moderate drop in scene graph recall (7.7 vs.\ 11.5 R@100). These results suggest that visual and textual inputs compliment to each other --- predicate prediction primarily relies on textual inputs while object prediction mainly depends on visual inputs.

\subsection{Fully-supervised Scene Graph Generation}
\label{SGG-in-domain}
We further evaluate our model for fully supervised SGG.

\noindent \textbf {Setup and Baselines}. To demonstrate the strength of our Transformer-based model, we present results on fully supervised SGG and compare to several latest methods~\cite{xu2017scene,zhang2017visual,tang2019learning,zellers2018neural}, following the standard protocal of training and testing on VG. Note that all models used the same object detector trained on VG and the same benchmark implementation provided by Tang \etal \cite{tang2020unbiased}.

\noindent \textbf{Results}. We report recall and mean recall for SGDet, SGCls, and PredCls in Table~\ref{tab:supervised-SGG}. The recalls of our model compare favorably to previous best results (SGDet: 36.3 vs.\ 36.9 R@100, SGCls: 40.8 vs.\ 41.4 R@100, PredCls: 67.4 vs.\ 68.1). More importantly, the mean recalls of our model are significantly higher than previous models across all evaluation protocols (SGDet: 8.7 vs.\ 6.9 R@100, SGCls: 11.1 vs.\ 8.7 R@100, PredCls: 19.5 vs.\ 16.1). Compared to recall, mean recall~\cite{chen2019knowledge,tang2019learning} better characterizes the performance on categories with fewer samples. These results suggest that our model is better at capturing those tail categories.

\subsection{Open-set Scene Graph Generation}
\label{SGG-openset}
Moving forward, we consider a challenging open-set setting for SGG, where the categories of target concepts (objects and predicates) are unknown during training. We believe this is the first result for open-set SGG. 

\noindent \textbf {Setup}. In this experiment, our model is trained on COCO Caption and evaluated on VG. During training, we parsed concept categories from captions, remove the low-frequency categories, and formed a vocabulary of 4273 objects and 677 predicates. This vocabulary was then used to train our model. At inference time, we first generated scene graphs using our vocabulary, and then matched the detected categories in our vocabulary to target concepts on VG (150 objects and 50 predicates) for evaluation.

\begin{figure}[t]
	\centering
	\includegraphics[width=0.95\linewidth]{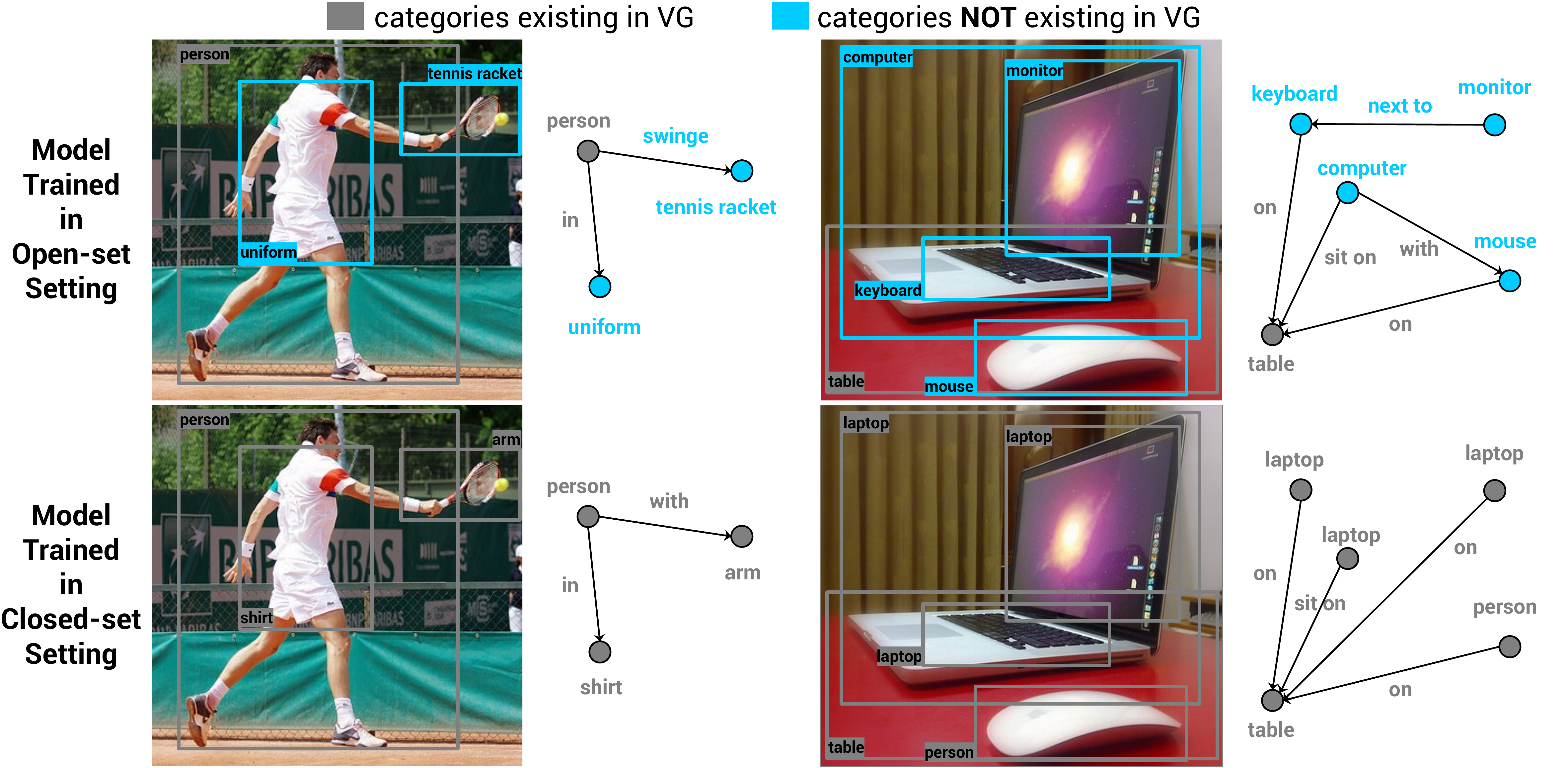}
    \caption{Qualitative results of our models (trained in open-set and closed-set settings) on VG test set for SGG.}
    \label{fig:openset_vis}
\end{figure}

\noindent \textbf {Results}. Table~\ref{tab:categories-SGG} compares the results of our models trained in closed-set and open-set settings using the same COCO caption dataset. The model trained in open-set setting has slightly better recall (4.8 vs.\ 4.5 R@100). Our open-set results are also comparable to VSPNet (supervised by unlocalized scene graphs on VG in a closed-set setting). We hypothesis that the open-set setting allows the model to learn from more concepts and thus benefits SGG. To verify this hypothesis, we plot the output scene graph from our models  trained on closed-set and open-set settings in Fig.\ \ref{fig:openset_vis}. Compared to our closed-set model, our open-set model detects more concepts outside VG (\eg ``swinge'', ``mouse'', ``keyboard''). Our results suggest an exciting avenue of large-scale training of open-set SGG using image captioning dataset such as CC.

\begin{table}[]
\centering
\resizebox{0.45\textwidth}{!}{%
\begin{tabular}{c|c|c|c|c|cc}
\hline
\multirow{2}{*}{Model} & \multirow{2}{*}{\#Objects} & \multirow{2}{*}{\#Predicates} & \multirow{2}{*}{\#Triplets} & \multirow{2}{*}{\#Images} & \multicolumn{2}{c}{SGDet} \\
 &  &  &  &  & R@50 & R@100 \\ \hline
Ours & 143 & 56 & 154K & 64K & 3.8 & 4.5 \\
Ours & 4273 & 677 & 758K & 105K & 4.1 & 4.8 \\\hline
\end{tabular}}\vspace{0.3em}
\caption{Results of open-set SGG. Evaluation is performed on VG with the vocabulary and model learned from COCO.}
\label{tab:categories-SGG}
\end{table}

%-------------------------------------------------------------------------
% conclusion
\section{Conclusion}
We proposed one of the first methods of learning to generate scene graphs from image-sentence pairs. Our key idea is to use off-the-shelf object detectors, so as to match detected object tags to parsed tokens from captions, thus creating ``pseudo'' labels for training. Further, we designed a Transformer-based model and demonstrated strong results across different levels of supervision. Our model learned from only image-sentence pairs, outperformed a state-of-the-art weakly supervised model trained by human-annotated unlocalized scene graphs. More importantly, we presented the first result for open-set scene graph generation. We hope our work points to exciting avenues of learning structured visual representation from natural language.\\
\noindent \textbf{Limitation and Future Work}. A main limitation of our method is the dependency on an object detector covering a wide range of concepts. We anticipate that our method will benefit from the development of open-vocabulary detectors~\cite{ye2019cap2det,zareian2020open}. Moreover, there is fundamental ambiguity during the label assignment step in learning from image-sentence pairs when there are multiple object instances of the same category. We hypothesis that contextual cues such as surrounding objects might help and will leave this as future work. Finally, we also plan to explore using scene graphs learned from image-sentence pairs for vision and language tasks (\eg VQA).\\
\\
\textbf{Acknowledgement}: YZ and YL acknowledge the support provided by the UW-Madison OVCRGE with funding from WARF. JS and CX were supported by the National Science Foundation (NSF) under Grant RI:1813709. The article solely reflects the opinions and conclusions of its authors but not the funding agency.

\bibliographystyle{ieee_fullname}
\bibliography{refs}

\newpage
\appendix
%\appendixpage
\addappheadtotoc
\begin{appendices}

In appendices, we provide additional implementation details, describe the domain gap among different datasets, and present detailed results of scene graph generation. We hope that this document will complement our main paper.

\section{Implementation Details}
We now present implementation details on triplet filtering (Sec.~\ref{section:training}) and our model architecture (Sec.~\ref{section:experiments}).\smallskip

\noindent \textbf{Triplet Filtering}.
We describe details for triplet filtering during an open-set setting. In the open-set setting, the subject, predicate and object categories in the target dataset are unknown during training. There is thus no limits on these categories. We simply removed the parsed concepts with low frequency. The frequency threshold was set to 3 and 10 for objects/subjects and predicates, respectively. At the end, we kept 4273-677 subject/object-predicate categories for COCO Caption dataset~\cite{chen2015microsoft}. These categories correspond to 758K triplet instances and 105K images. These images and triplet instances were used for model training in the open-set setting.\smallskip

\noindent \textbf{Model Architecture}.
Table~\ref{tab:model-arch} lists the architecture of our Triplet Transformer. Our model consists of the visual embedder, the textual embedder, the multi-layer Transformer encoder, and the classification heads. In the Transformer encoder, each self-attention layer~\cite{transformer} takes the token embeddings as inputs, and outputs a contextualized embedding for each token. We used an input/output dimension of 768-D with 12 attention heads each with 64 dimensions.

\section{Domain Gap among Different Datasets}
% data distribution / image and caption samples
Further, we study the domain gap among the datasets considered in the paper (Conceptual Caption~\cite{sharma2018conceptual}, COCO Caption~\cite{chen2015microsoft}, Visual Genome~\cite{krishna2017visual}). Our main results are obtained by training our model on Conceptual Caption and COCO, and evaluating the trained model on Visual Genome. Our observation is that the gap between Conceptual Caption and COCO Caption vs.\ Visual Genome is large and thus our task is very challenging.

\begin{figure*}
	\centering
	\includegraphics[width=0.88\linewidth]{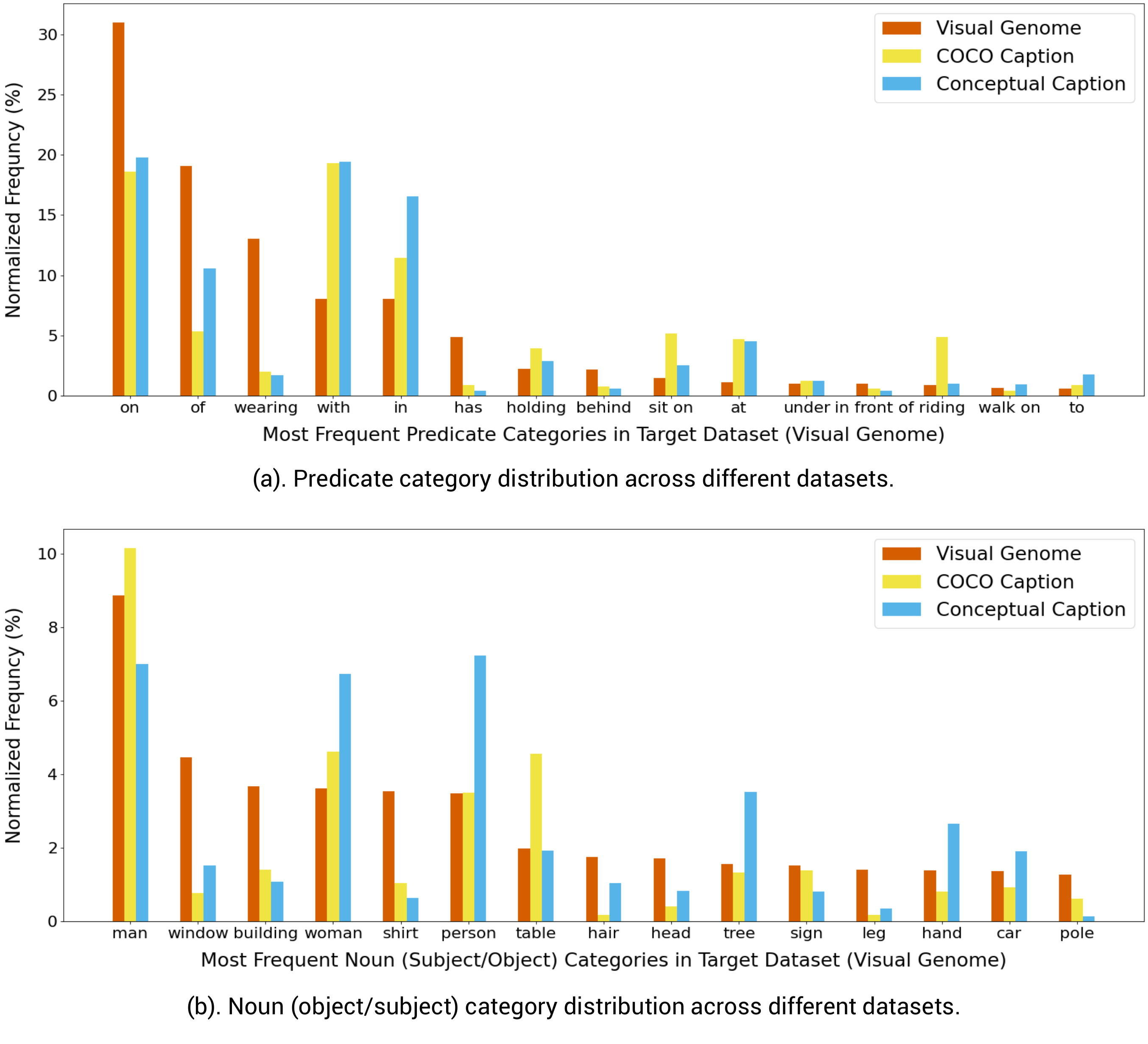} 
    \caption{The distribution of (a) the predicate categories and (b) the noun (subject/object) categories parsed from different image caption datasets (Visual Genome, COCO Caption and Conceptual Caption). Y-axis represents the category frequency normalized in respective dataset. X-axis indexes the most common categories in the target dataset (Visual Genome). These three datasets have very different distributions with a large domain gap.}
    \label{fig:data-dist}
\end{figure*}

To verify our observation, we plot the distribution of the most frequent predicate categories (a) and noun (subject/object) categories (b) across three datasets in Figure~\ref{fig:data-dist}. These categories are parsed from image captions and matched to a target dictionary defined on Visual Genome. The top 15 most frequent noun/predicate categories are displayed. The domain gap between different datasets can be clearly observed in Figure~\ref{fig:data-dist}. For example, the predicate category ``wearing'' is very common in Visual Genome dataset, yet quite rare in COCO Caption and Conceptual Caption. As described in Sec.~\ref{section:training}, we applied the weighted loss during training to bridge the domain gap. According to our experiments in Sec.~\ref{section:ablation}, the weighted loss can help to improve the performance of scene graph generation.

% table for per-category recall
\begin{table*}
\centering
\resizebox{0.98\textwidth}{!}{%
\begin{tabular}{c|c|c|ccccccccccccccc}
\hline
Models & \multicolumn{1}{l|}{Supervision} & R@100 & on & has & in & of & wearing & near & with & above & holding & behind & sit on & in front of & at & riding & eating \\ \hline
Ours & CC+COCO & 7.0 & 7.4 & 1.0 & 3.8 & 6.2 & 24.4 & 8.4 & 7.4 & 0.4 & 5.3 & 0.4 & 3.5 & 2.0 & 8.4 & 3.7 & 17.9 \\
Ours & VG-Weak & 11.5 & 12.3 & 14.5 & 4.5 & 6.9 & 27.9 & 6.8 & 0.9 & 0.7 & 19.1 & 12.7 & 7.9 & 3.5 & 18.9 & 18.4 & 10.6 \\
Ours & VG-Full & 15.3 & 16.8 & 15.8 & 7.6 & 7.5 & 38.4 & 10.5 & 1.0 & 1.4 & 23.2 & 13.6 & 6.8 & 2.6 & 9.5 & 22.8 & 8.9 \\ \hline
\end{tabular}}
\caption{Per-category recall@100 for scene graph generation. Each row corresponds to one of the models in Table~\ref{tab:weakly-SGG} of our main paper. CC+COCO represents the model trained by the image captions from Conceptual Caption (CC) and COCO Caption (COCO). VG-Weak represents the model trained by the unlocalized scene graphs in Visual Genome. VG-Full denotes the model trained by the localized scene graphs in Visual Genome. We list the overall recall as well as the recalls for most frequent predicates.}
\label{tab:per-category-recall}
\end{table*}

\section{Further Analysis of Language Supervised Scene Graph Generation}
% break down of the results

We present additional details for our main results on language supervised scene graph generation, shown in Table 2 of our main paper. Specifically, we provide a breakdown of our results, and compare our models obtained by different levels of supervision.\smallskip 

\noindent \textbf{Setup}. We present additional details in the setting of SGDet R@100, shown in Table~\ref{tab:weakly-SGG} of our main paper. We show the per-category recalls for 15 most common predicate categories. These categories are selected by their frequency in the scene graph annotation of Visual Genome. The per-category recalls are calculated in the same way as the mean Recall (mR) except that we present the individual recall before averaging over all categories.\smallskip

\noindent \textbf{Results}. Table~\ref{tab:per-category-recall} shows the per-category recalls for our models trained by different supervisory signals, including language supervised, weakly supervised and fully supervised. Each row in Table~\ref{tab:per-category-recall} corresponds to one of the models in Table~\ref{tab:weakly-SGG} of our main paper (\eg, CC+COCO represents our model trained by image descriptions from Conceptual Caption and COCO Caption). The per-category recalls vary drastically among different supervision settings. For example, our model trained by the localized scene graph labels (fully supervised) achieves the highest recall (16.8) for the category ``on'', while the model trained by image captions (language supervised) obtains the lowest recall (7.4). On the other hand, our language supervised model largely outperforms fully supervised and weakly supervised models on several categories, such as ``with'' (7.4 vs.\ 0.9 and 1.0) and ``eating'' (17.9 vs.\ 10.6 and 8.9). We conjecture that the performance difference is again produced by the domain gap of the supervisory signals (\ie datasets). For example, ``eating'' is more frequent on Conceptual Caption and COCO Caption than on Visual Genome. It will be an exciting direction to investigate how to further bridge this domain gap.

\cleardoublepage

% table for model architecture
\begin{table*}[]
\centering
\resizebox{0.98\textwidth}{!}{%
\begin{tabular}{|c|c|c|c|c|c|c|}
\hline
ID & Module & Input ID & Layer Type & Weights Size & Output Size & Comments \\ \hline
1 & \multirow{3}{*}{Input} & - & Region Features & - & (N, 1536) & \multirow{3}{*}{\begin{tabular}[c]{@{}c@{}}Obtained from\\ the off-the-shelf\\ object detector\end{tabular}} \\ \cline{1-1} \cline{3-6}
2 &  & - & Region Boxes & - & (N, 7) &  \\ \cline{1-1} \cline{3-6}
3 &  & - & Region Tags & - & (N) &  \\ \hline
4 & \multirow{7}{*}{\begin{tabular}[c]{@{}c@{}}Visual\\ Embedder\end{tabular}} & 1 & FC & (1536, 768) & (N, 768) &  \\ \cline{1-1} \cline{3-7} 
5 &  & 4 & LN & (768) & (N, 768) &  \\ \cline{1-1} \cline{3-7} 
6 &  & 2 & FC & (7, 768) & (N, 768) &  \\ \cline{1-1} \cline{3-7} 
7 &  & 6 & LN & (768) & (N, 768) &  \\ \cline{1-1} \cline{3-7} 
8 &  & - & Type Embedding & (3, 768) & (N, 768) & Region type: subject vs. object vs. context \\ \cline{1-1} \cline{3-7} 
9 &  & 5+7+8 & LN & (768) & (N, 768) &  \\ \cline{1-1} \cline{3-7} 
10 &  & 9 & Dropout & - & (N, 768) & p=0.1 \\ \hline
11 & \multirow{5}{*}{\begin{tabular}[c]{@{}c@{}}Textual\\ Embedder\end{tabular}} & 3 & Word Embedding & (604, 200) & (4, 200) & Text tokens: [subject tag, MASK, object tag, SEP] \\ \cline{1-1} \cline{3-7} 
12 &  & 11 & FC & (200, 768) & (4, 768) &  \\ \cline{1-1} \cline{3-7} 
13 &  & - & Position Embedding & (4, 768) & (4, 768) & Embedding for text token position \\ \cline{1-1} \cline{3-7} 
14 &  & 12+13 & LN & (768) & (4, 768) &  \\ \cline{1-1} \cline{3-7} 
15 &  & 14 & Dropout & - & (4, 768) & p=0.1 \\ \hline
16 & \multirow{10}{*}{\begin{tabular}[c]{@{}c@{}}Trainsformer\\ Encoder\\ Layer 1\end{tabular}} & [10, 15] & Concatenation & - & (N+4, 768) & Concatenate the visual and textual features \\ \cline{1-1} \cline{3-7} 
17 &  & 16 & Self-attention & (768, 64, 3, 12) & (N+4, 768) & Multi-head self-attention~\cite{transformer} \\ \cline{1-1} \cline{3-7} 
18 &  & 17 & FC & (768, 768) & (N+4, 768) &  \\ \cline{1-1} \cline{3-7} 
19 &  & 18 & Dropout & - & (N+4, 768) & p=0.1 \\ \cline{1-1} \cline{3-7} 
20 &  & 16+19 & LN & (768) & (N+4, 768) & Residual connection followed by LN \\ \cline{1-1} \cline{3-7} 
21 &  & 20 & FC & (768, 3072) & (N+4, 3072) &  \\ \cline{1-1} \cline{3-7} 
22 &  & 21 & GELU & - & (N+4, 3072) &  \\ \cline{1-1} \cline{3-7} 
23 &  & 22 & FC & (3072, 768) & (N+4, 768) &  \\ \cline{1-1} \cline{3-7} 
24 &  & 23 & Dropout & - & (N+4, 768) & p=0.1 \\ \cline{1-1} \cline{3-7} 
25 &  & 20+24 & LN & (768) & (N+4, 768) & Residual connection followed by LN \\ \hline
26 & \multirow{9}{*}{\begin{tabular}[c]{@{}c@{}}Transformer\\ Encoder\\ Layer 2\end{tabular}} & 25 & Self-attention & (768, 64, 3, 12) & (N+4, 768) & Multi-head self-attention~\cite{transformer} \\ \cline{1-1} \cline{3-7} 
27 &  & 26 & FC & (768, 768) & (N+4, 768) &  \\ \cline{1-1} \cline{3-7} 
28 &  & 27 & Dropout & - & (N+4, 768) & p=0.1 \\ \cline{1-1} \cline{3-7} 
29 &  & 25+28 & LN & (768) & (N+4, 768) & Residual connection followed by LN \\ \cline{1-1} \cline{3-7} 
30 &  & 29 & FC & (768, 3072) & (N+4, 3072) &  \\ \cline{1-1} \cline{3-7} 
31 &  & 30 & GELU & - & (N+4, 3072) &  \\ \cline{1-1} \cline{3-7} 
32 &  & 31 & FC & (3072, 768) & (N+4, 768) &  \\ \cline{1-1} \cline{3-7} 
33 &  & 32 & Dropout & - & (N+4, 768) & p=0.1 \\ \cline{1-1} \cline{3-7} 
34 &  & 29+33 & LN & (768) & (N+4, 768) & Residual connection followed by LN \\ \hline
35 & \multirow{19}{*}{\begin{tabular}[c]{@{}c@{}}Classification\\ Heads\end{tabular}} & 34 & Indexing & - & (1, 768) & Subject visual embedding \\ \cline{1-1} \cline{3-7} 
36 &  & 34 & Indexing & - & (1, 768) & Object visual embedding \\ \cline{1-1} \cline{3-7} 
37 &  & 34 & Indexing & - & (1, 768) & Subject textual embedding \\ \cline{1-1} \cline{3-7} 
38 &  & 34 & Indexing & - & (1, 768) & Object textual embedding \\ \cline{1-1} \cline{3-7} 
39 &  & 34 & Indexing & - & (1, 768) & Predicate embedding \\ \cline{1-1} \cline{3-7} 
40 &  & 11 & Indexing & - & (1, 200) & Subject word embedding \\ \cline{1-1} \cline{3-7} 
41 &  & 11 & Indexing & - & (1, 200) & Object word embedding \\ \cline{1-1} \cline{3-7} 
42 &  & 40 & \multirow{2}{*}{FC} & \multirow{2}{*}{(200, 768)} & (1, 768) & \multirow{2}{*}{} \\ \cline{1-1} \cline{3-3} \cline{6-6}
43 &  & 41 &  &  & (1, 768) &  \\ \cline{1-1} \cline{3-7} 
44 &  & 35+42 & \multirow{2}{*}{FC, ReLU, FC} & \multirow{2}{*}{\begin{tabular}[c]{@{}c@{}}(768,768), - ,\\ (768,151)\end{tabular}} & (1, 151) & Subject logits \\ \cline{1-1} \cline{3-3} \cline{6-7} 
45 &  & 36+43 &  &  & (1, 151) & Object logits \\ \cline{1-1} \cline{3-7} 
46 &  & 44 & Softmax & - & (1, 151) & Used for subject cross-entropy loss \\ \cline{1-1} \cline{3-7} 
47 &  & 45 & Softmax & - & (1, 151) & Used for object cross-entropy loss \\ \cline{1-1} \cline{3-7} 
48 &  & 35 & FC & (768, 768) & (1, 768) &  \\ \cline{1-1} \cline{3-7} 
49 &  & 36 & FC & (768, 768) & (1, 768) &  \\ \cline{1-1} \cline{3-7} 
50 &  & 37 & FC & (768, 768) & (1, 768) &  \\ \cline{1-1} \cline{3-7} 
51 &  & 38 & FC & (768, 768) & (1, 768) &  \\ \cline{1-1} \cline{3-7} 
52 &  & \begin{tabular}[c]{@{}c@{}}39+\\ 48+49+\\ 50+51\end{tabular} & FC, ReLU, FC & \begin{tabular}[c]{@{}c@{}}(768,768), - ,\\ (768,151)\end{tabular} & (1, 51) & Predicate logits \\ \cline{1-1} \cline{3-7} 
53 &  & 52 & Softmax & - & (1, 51) & Used for predicate cross-entropy loss \\ \hline
\end{tabular}}
\caption{Model architecture of our proposed Triplet Transformer. For each input image, $N=36$ detected regions from the off-the-shelf object detector are used as our model inputs. Within the Input ID column, ``+'' means summing up the input features, ``[]'' means concatenating the input features. Within the Layer Type column, ``FC'' denotes the fully-connected layer, ``LN'' denotes the Layer Normalization~\cite{ba2016layer}, ``GELU'' represents the GELU activation function~\cite{gelu}, and ``Indexing'' means extracting the features corresponding to a particular slot (\eg, visual or textual embeddings of subject, predicate and object.).}
\label{tab:model-arch}
\end{table*}

\cleardoublepage

\end{appendices}

\end{document}